\documentclass{article}

\usepackage[english]{babel}

\usepackage[letterpaper,top=2cm,bottom=2cm,left=3cm,right=3cm,marginparwidth=1.75cm]{geometry}
\usepackage{endnotes}
\let\footnote=\endnote
\usepackage{amsmath}
\usepackage{graphicx}
\usepackage{hyperref}
\hypersetup{colorlinks=true, allcolors=blue}
\begin{document}
\title{Kratt: Developing an Automatic Subject Indexing Tool for The National Library of Estonia}
\author{Marit Asula, Jane Makke, Linda Freienthal, Hele-Andra Kuulmets \& Raul Sirel}
\date{28 May 2021}

\maketitle
This is a preprint version of Marit Asula, Jane Makke, Linda Freienthal, Hele-Andra Kuulmets \& Raul Sirel (2021) Kratt: Developing an Automatic Subject Indexing Tool for the National Library of Estonia, Cataloging \& Classification Quarterly, 59:8, 775-793, DOI: 10.1080/01639374.2021.1998283
\begin{abstract}
Manual subject indexing in libraries is a time-consuming and costly process and the quality of the assigned subjects is affected by the cataloguer's knowledge on the specific topics contained in the book. Trying to solve these issues, we exploited the opportunities arising from artificial intelligence to develop Kratt: a prototype of an automatic subject indexing tool. Kratt is able to subject index a book independent of its extent and genre with a set of keywords present in the Estonian Subject Thesaurus. It takes Kratt approximately 1 minute to subject index a book, outperforming humans 10-15 times. Although the resulting keywords were not considered satisfactory by the cataloguers, the ratings of a small sample of regular library users showed more promise. We also argue that the results can be enhanced by including a bigger corpus for training the model and applying more careful preprocessing techniques.

\end{abstract}

\section{Keywords}
national libraries, automated subject indexing, machine learning, natural language processing, cataloguing

\section{Introduction}
As a national bibliographic agency, the National Library of Estonia is responsible for the registration of the publications issued in Estonia or outside Estonia by Estonians. Bibliographic descriptions are primarily definitive containing all the mandatory elements set out in the ISBD. Normally, the cataloguing process also includes the subject indexing for which Estonian Subject Thesaurus (Eesti Märksõnastik, EMS)\footnote{Estonian Subject Thesaurus \url{https://ems.elnet.ee/}} is used along with UDC Summary for classifying the resources. EMS was launched in 2009, it currently contains about 61 000 terms, among which there are approximately 40 000 preferred terms and 21 000 non-preferred terms. The terms in EMS are in Estonian, however, an English translation is given for each term. 

According to some authors, the cataloguing process is considered to be time-consuming and expensive in the library work\footnote{ Kont, K.-R., 2015. “How Much Does It Cost to Catalog a Document? A Case Study in Estonian University Libraries”. Cataloging \& Classification Quarterly, 53:7 (2014), 825-50, DOI: 10.1080/01639374.2015.1020463 }: cataloguing is manual work and it is estimated that a cataloger is able to describe around 3-4 books per hour\footnote{Pokorny, J. “Automatic Subject Indexing and Classification Using Text Recognition and Computer-Based Analysis of Tables of Contents”. Toronto, ELPUB, 2018. DOI: 10.4000/proceedings.elpub.2018.19}. At the National Library of Estonia, it is estimated that the cataloguer should produce 10 national bibliographic records\footnote{According to the Estonian National Bibliography, the annual production of publications is approximately 11 000 items (both print and e-publications).} in a day depending on the material type. 

Subject access data is usually created by a cataloguer who discovers the topics of the book and performs a content analysis. Eventually, keywords expressing the subject of the resource will be generated. It is estimated that the subject indexing and classifying of resources takes approximately 15 minutes depending on the material, topic and the level of experience of the librarian. Indeed, one of the problems with subject description performed by catalogers is the limited ability to understand and describe the subject. The quality of description depends on the intellectual assumptions of a cataloger and is affected by some amount of subjectivity\footnote{Cloete, L. M.; Snyman, R.; Cronje, J.C., “Training Cataloguing Students Using a Mix of Media and Technologies”. Aslib Proceedings, 55:4 (2003), 223–233. DOI: 10.1108/00012530310486584}. Pokorny argues: “When processing scientific books, a cataloger who is not an expert in a given discipline is not able to correctly and precisely understand and describe the topics contained in the book.”\footnote{Pokorny, “Autmatic Subject Indexing”}

Wishing to address the above-mentioned issues (speed, subjectivity, cost and quality), the National Library of Estonia decided to test automating the process of subject indexing with the help of machine learning and text mining tools. Since Estonian Government is actively supporting the development of artificial intelligence\footnote{National artificial intelligence strategy for 2019-2021 https://en.kratid.ee/ } to reduce costs, raise quality and save time in the public sector, the Library found financial support for the project from the government programs. It also sought help from the private sector to find relevant knowledge and experience.  

The project was initiated in 2019 and it took 6-7 months to build a prototype of the Kratt\footnote{In Estonian mythology, Kratt is a magical creature. Essentially, Kratt was a servant built from hay or old household items by its master who then had to cede three drops of blood for the devil to bring life to the kratt. The kratt was notable for doing everything the master ordered. Therefore, the Estonian government uses this character as a metaphor for AI and its complexities} (it is a common name for AI applications in Estonia) for automated subject indexing of books in Estonian language. Library’s goal was to test: (a) whether the automation of the process would be possible, (b) if it might save time and money, (c) if Kratt could help to raise the quality, and (d) if Kratt could be integrated into the daily cataloguing workflows. 

\section{Related work}

Several national libraries have over the years reported their attempts to exploit text mining and machine learning methods in order to automate cataloguing and indexing tasks and reduce the amount of human workload needed. 

In general, there are two options for libraries to approach the issue. Either they build their own solution from scratch or purchase the software from the market. The later option for example has been used by Deutsche Nationalbibliothek\footnote{Junger, U., 2014. “Can Indexing Be Automated? The Example of the Deutsche Nationalbibliothek”. Cataloging \& Classification Quarterly, 52:1, 102-9. DOI: 10.1080/01639374.2013.854127}. The subject indexing tool that they evaluated uses unsupervised methods to extract terms and later matches them to the controlled vocabulary. The results of their experiments, which focused on online publications (mostly doctoral theses) were not considered satisfactory, mainly because of the low precision of assigned subjects. 

A  different approach was chosen by the National Library of Finland who developed its own tool for subject indexing and text classification.\footnote{Suominen, O., 2019. “Annif: DIY Automated Subject Indexing Using Multiple Algorithms”. LIBER Quarterly, 29 (1), 1–25. DOI: 10.18352/lq.10285} The tool, called Annif, is built on top of existing open-source algorithms, allowing users to choose from multiple unsupervised and supervised algorithms, including ensemble methods. Their experiments, conducted on more or less academic articles, Master’s and Doctoral theses, question-answer pairs of any topic and a regional newspaper, showed that ensemble methods perform better than individual methods. Annif performed best on theses with an average f1-score of 0.46 and worst on newspaper articles with an average f1-score of 0.28. It is also reported that in The University of Jyväskylä, where Annif was adapted, approximately one half of the subjects suggested were selected as final subjects by students who were uploading their Master’s theses to the repository. Librarians who were reviewing uploads selected 53\% of the same suggestions.

\section{Description of the Training Data}

\begin{figure}[h]
\centering
\includegraphics[width=\textwidth]%,height=\textheight,keepaspectratio]
{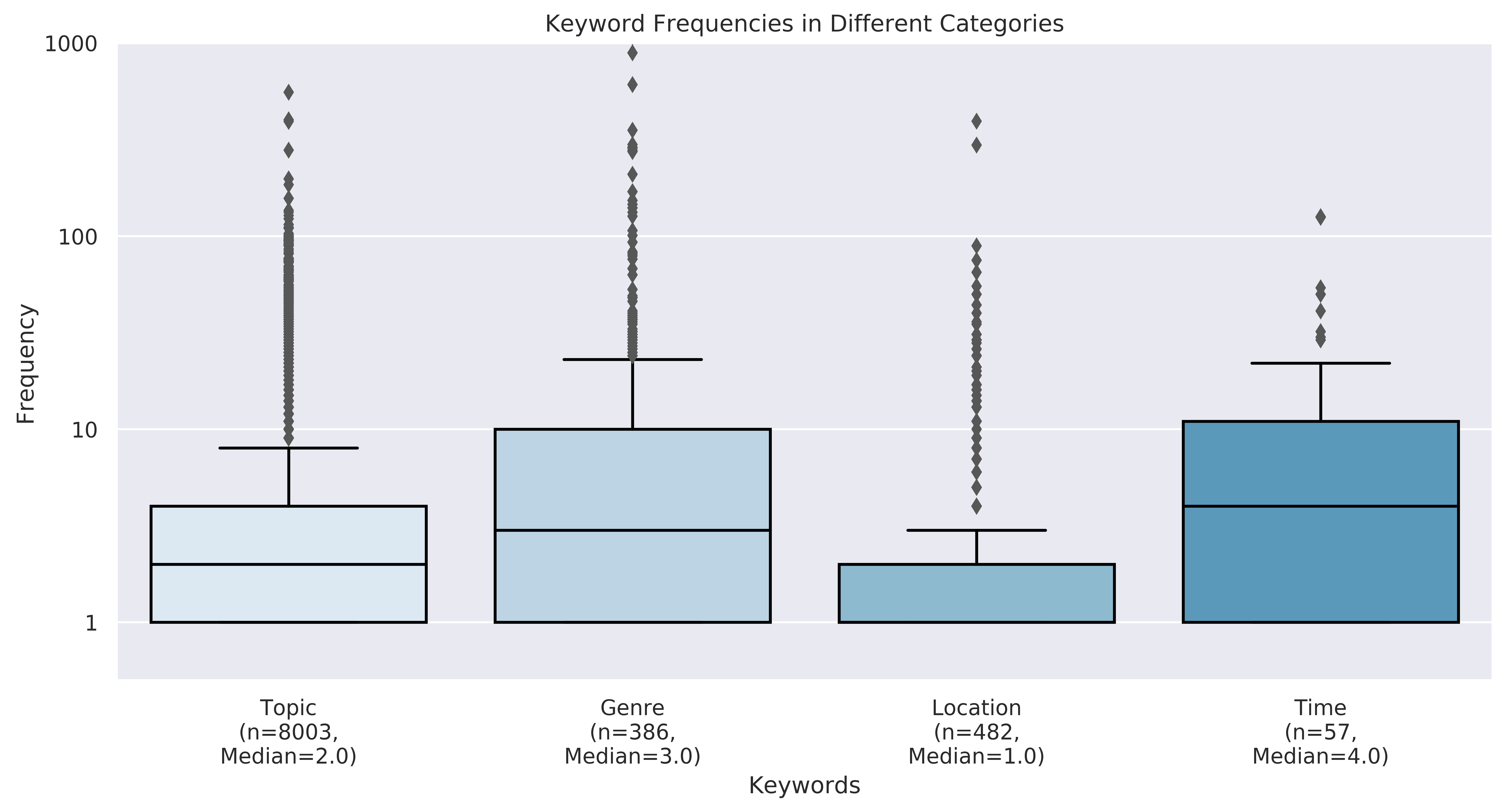}
\caption{The frequency distributions of unique keywords in the training set. Each boxplot represents the distribution of the keywords in a specific category. }
\label{figure_1_kw_frequency_distribution}
\end{figure}

The National Library of Estonia (NLE) provided 7668 publically available books with corresponding subject indices for developing the prototype. All the books used for developing the prototype are available in NLE’s digital archive DIGAR\footnote{DIGAR stores online publications, print files and digitised copies of publications. \url{https://www.digar.ee/arhiiv/en}} and the corresponding subject indices in the Estonian National Bibliography (ERB)\footnote{ERB  registers all publications issued in Estonia, as well as Estonian-language publications, works by Estonian authors and their translations, and foreign-language publications about Estonia and Estonians issued abroad. \url{https://erb.nlib.ee/?lang=eng}}. The books were in various languages and consisted of a wide variety of forms, including dissertations, reports, brochures, manuals, textbooks, collections of articles, transcripts, dictionaries, novels, short stories etc. Each book had been subject indexed by a professional cataloguer and the total number of unique preferred terms (also referred to as “subject indices”, “labels” or “keywords” in the subsequent text) was 10 098. Each preferred term belonged to one of the seven major categories: genre and form (e.g. "fiction", "memorial"), time (e.g. "21st century", "2020"), location (e.g. "Latvia", "London"), topic (e.g. "sewing", “economy”), person (“Barack Obama”, “Herman Hesse”), collective/organization (e.g. “The European Union”, “The University of Tartu”), and temporary collective or event (e.g. “The European Capital of Culture”, “Black Nights Film Festival”). As categories “person”, “collective”, “temporary collective or event” are not based on EMS or any other thesaurus and would have needed additional preprocessing, e.g. merging differently written preferred terms referring to the same entity, we excluded the labels belonging to these categories from the prototype. After removing the aforementioned keywords, the total number of labels was reduced to 8928. As EMS contains about 40 0000 preferred terms, it reveals the first shortcoming of the training data: only 22\% of all the possible labels were included. This means that supervised machine learning methods would be able to predict only the same subset of labels excluding 78\% of all the possible labels. Furthermore, 8928 labels might constitute only one fifth of all the preferred terms in EMS, but it is still a very large set of possible targets to consider while constructing the keyword assignment models. Another difficulty arising from the data was a very sparse distribution of unique labels: the median frequency of unique labels was 2, with most of the labels occurring only 1-4 times in the whole set (Figure \ref{figure_1_kw_frequency_distribution}). This is an extremely small number considering the fact that most machine learning-based classification methods require hundreds of examples to accurately learn the features corresponding to each label. If we set the minimum number of required examples to 50 - a relatively low bar to cross, only 111 labels would have exceeded the threshold. 

There is no specific limit set on the number of labels per one document in the library’s subject indexing methodology, but the set of labels should be sufficiently large to provide an adequate overview of the book’s content and key topics, yet as succinct and compact as possible. The number of labels per document in the training set varies from 1 to 35 with an average number of 5. 

86.5\% of the books in the training set were in Estonian, 9.8\% in English, 2.5\% in Russian and 1.2\% in other languages, including Finnish, German, French, Latvian, Swedish etc. 

The extent of the books varied from 1 to 1936 pages, with a mean number of 86 and a median number of 52.

The subject index contained the publishing dates only for a specific, often translated, edition of the books and did not have direct information of the original publishing dates. However, it did include the author's date of birth, so we could  approximate the original publishing dates based on this data. It was necessary to obtain this information as the disparity between two-word distributions grows with their temporal difference and as word frequencies influence the statistical prediction models, this could thus have an effect on the results\footnote{Jatowt, A.; Tanaka, K., 2012. “Large Scale Analysis of Changes in English Vocabulary over Recent Time”. CIKM '12: Proceedings of the 21st ACM international conference on Information and knowledge management, 2523-26. DOI: 10.1145/2396761.2398682}. For example, if most of the training data is published after the year 2000, it would be safe to assume that the model would perform poorer on books originally published in the 15th century. The oldest author in the training data was Niccolò Machiavelli (born in 1469), followed by Wilhelm Christian Friebe (born in 1762) and Carl Friedrich von Ledebour (born in 1785). However, all the other authors present in the training set were born after the year 1800 and the majority (93.5\%) in the 20th century. Assuming that the published books are usually written by authors who are at least 17 years old, we can derive that  93\% of the books in the training set were originally published after the year 1917, i.e. in the last 100 years.

As the performance and the results of machine learning methods strongly depend on the underlying data, it is important to be aware of the limitations arising from it. Most of the training documents were in Estonian (86.5\%) and 93\% of the books were published in the last 100 years. It should also be taken into account that the distribution of the genres and forms of the books in the public domain might differ from the distribution of genres and forms of the copyrighted books. 

\section{Constructing the Prediction Models}

Keyword assignment methods can be roughly divided into two parts: 1) keyword extraction, where keywords are chosen from words that are explicitly mentioned in the original text and 2) keyword assignment, where keywords are chosen from a controlled vocabulary or taxonomy. As all the chosen keywords should be present in EMS, the latter technique is more suitable for following the methodology used for subject indexing by the National Library of Estonia. 

As the training data is labelled, it could otherwise be an ideal input for supervised machine learning methods, but we would first need to solve the following problems: 
\begin{enumerate}
    \item How to make the machine learning methods that require hundreds of examples to acquire the necessary information if the average number of examples per label are two? 
    \item How to time-efficiently predict a label set of size 5-10 from more than 8000 possible label candidates?
\end{enumerate}

To overcome the issue of the low number of training examples per label, we split each book into pages and linked all the labels in the book’s subject index with all of its pages. For instance, a book with 456 pages would have contributed 456 examples to every label it had been assigned. It should be noted, however, that this approach had a drawback: some of the books in the training set consisted of multiple unrelated sections, e.g. article collections and short story collections. Moreover, each label assigned to the book might have not represented every page of the book equally, even if the content distribution in the book was more or less uniform. This means that some of the labels for some of the pages might have been inaccurate, but we expected that the noise arising from these mismatches would not have a noticeable impact on the grand scale.

After splitting the documents and treating every page as a separate instance, the number of examples per keyword increased considerably, with the median frequency of 393. We set the minimum number of examples for each classifier model to 50 to guarantee sufficient number of examples for each keyword and obtained the final label set of size 8003.

The text from each page is thereafter extracted with Apache TIKA\footnote{\url{https://tika.apache.org/}} - an open-source tool supporting text extraction from a wide variety of formats, including images requiring optical character recognition. 

Each text is then passed through quality control by feeding it to a Hidden Markov model based on the distribution of character sequences to distinguish unsuccessful text extraction (texts consisting of meaningless character sequences like “AXwQkKSj4G”) from correctly extracted texts. 

The texts passing the quality check are processed with a multilingual preprocessing tool Texta MLP, which uses EstNLTK\footnote{EstNLT is an open-source tool for Estonian natural language preprocessing. \url{https://estnltk.github.io/estnltk/1.4.1/}} and spaCy\footnote{spaCy is a free, open-source library for advanced Natural Language Processing (NLP) in Python. \url{https://spacy.io/}} for identifying the language of the text, lemmatizing it and extracting part-of-speech (POS) tags. Lemmatization is a process of converting words into dictionary forms (running, ran and runs into run), which helps to reduce the size of the vocabulary and therefore provide more succinct and precise features for the machine learning models. POS tags (e.g. “noun”, “verb”, “adverb” etc) provide morphological information about the words based on their context and definition and have been successfully used to improve the results of classifying text genres. \footnote{Feldman, S.; Marin, M. A.; Ostendorf, M; Gupta, M. R., 2009. “Part-of-speech Histograms for Genre Classification of Text”, Proceedings of the IEEE International Conference on Acoustics, Speech, and Signal Processing, ICASSP 2009, 19-24 April 2009, Taipei, Taiwan. DOI: 10.1109/ICASSP.2009.4960700} 

The extracted lemmas and POS tags were used as input features for 8003 binary Logistic Regression classifiers - one for every label in the training set. 

\begin{figure}[h]
\centering
\includegraphics[width=\textwidth]%,height=\textheight,keepaspectratio]
{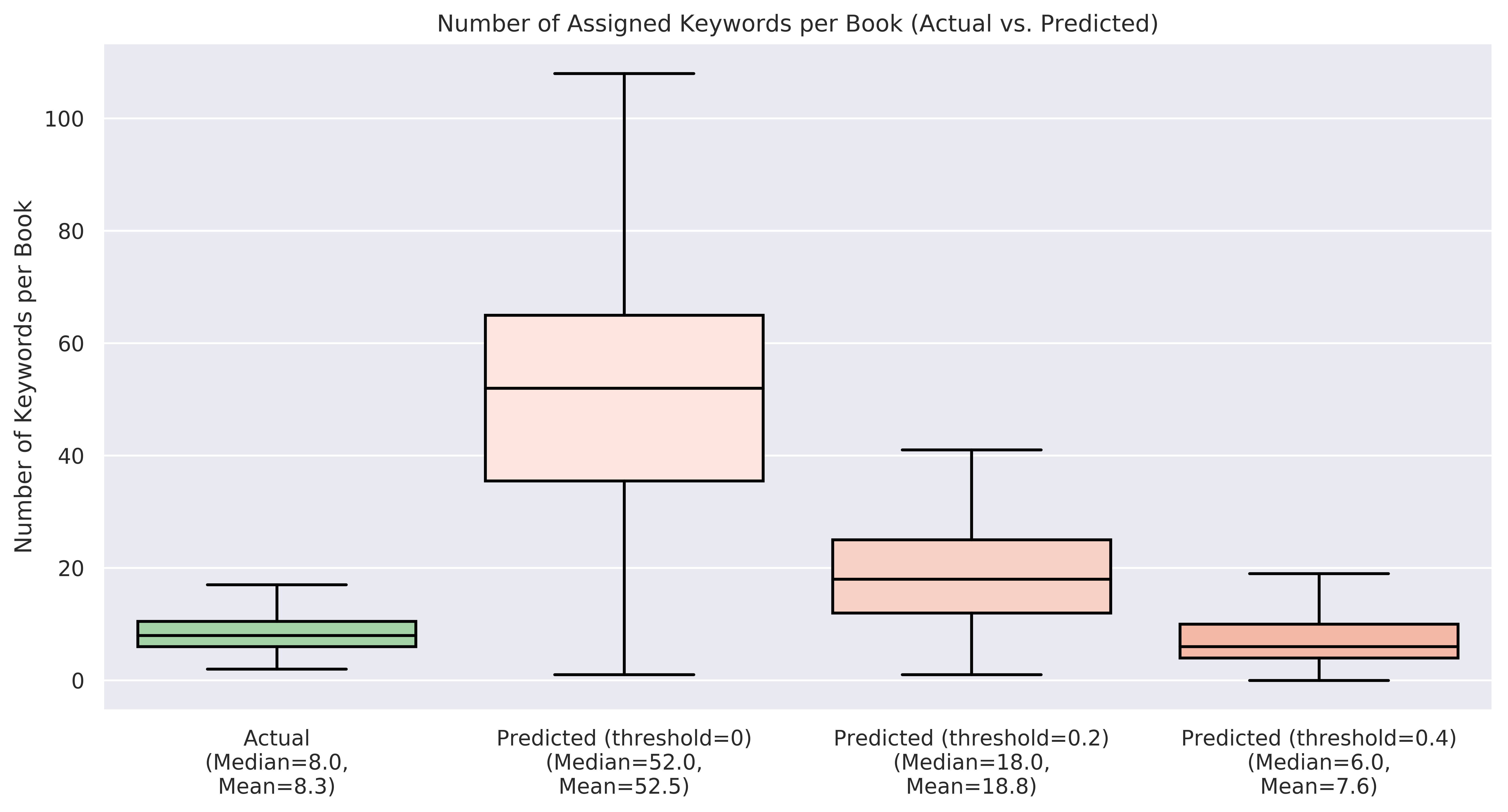}
\caption{The number of keywords assigned to the books by a) the cataloguers and b) the automatic subject indexing tool on different frequency thresholds.}
\label{figure_2_keywords_per_book}
\end{figure}

To time-efficiency predict the labels from 8003 possible candidates, we used Hybrid Tagger - a tool aimed at extreme multi-label classification tasks. Hybrid Tagger is part of TEXTA Toolkit\footnote{Texta Toolkit is an open-source framework for building and executing machine learning pipelines and analysing textual content.} and it enables to reduce the set of candidate tags significantly by comparing the input document to the other documents from the same domain indexed in Elasticsearch.\footnote{Vaik, K.; Asula, M.; Sirel, R., 2020. “Hybrid Tagger – An Industry-driven Solution for Extreme Multi-label Text”, Proceedings of the LREC2020 Industry Track, 26–30.DOI: 10.5281/zenodo.4306169} The set of candidate tags is constructed of the top n most frequent tags present in the m most similar documents. The binary classifiers corresponding to each candidate tag are then applied to the input document and tags with positive classification results are returned as the predicted labels. We used configuration n = 10 and m = 20, which means that each input document has at most 20 candidate tags reducing the total number of candidates 8003/20 = 400.5 times and thus making the prediction process notably faster. The workflow of Hybrid Tagger is depicted on Figure \ref{figure_2_keywords_per_book}.

If the book to annotate is 500 pages long, it might not be necessary to use all the pages for the subject indexing process as a) the time it takes the automatic subject indexing tool to predict the keywords grows with the number of pages to process and b) the results stabilize after a certain page limit or even get worse if the number of irrelevant labels grows. To determine how many pages of the book we should pass to the automatic subject indexing tool, we analysed how the recall scores changed while incrementing the number of pages. We ignored f1-score-based performance at this stage as we used another method addressed in the next paragraph for specifically optimizing precision. As the number of pages it takes for the results to converge might differ depending on the extent of the book, we constructed 5 classes based on the extent and selected 20 random examples for each class.  On average, it took 34 pages to fully converge, but only 6.6 to reach 90\%. Moreover, the 90\% was always reached after using at most 10 segments regardless of the book’s extent. The results for each extent class are presented in Table \ref{tab:table1}. We have estimated that processing and annotating one page takes about 6 seconds, although the time may vary depending on the available computational power of the machine hosting the subject indexing tool. This means that it takes on average 3 additional minutes to fully converge after reaching the 90\% (34.0*6 - 6.6*6 = 168 seconds). Based on these results, we decided to set the default number of pages to annotate to 10 as it is sufficient to reach at least 90\% of highest possible recall on average, yet ensures a fairly time-efficient subject indexing process. However, we decided to leave the user an option to modify it. 

\begin{table}
\centering
\begin{tabular}{p{.15\linewidth}|p{.40\linewidth}|p{.35\linewidth}}
\textbf{Extent} & \textbf{Average number of pages to reach 90\% of the maximum recall} & \textbf{Average number of pages to reach the maximum recall}\\\hline
1-49 pages & 10 & 14 \\
50-99 pages & 8 & 30 \\
100-299 pages & 4 & 39 \\
300-499 pages & 5 & 42 \\
>= 500 pages & 6 & 45 \\
\textbf{Average}& 6.6 & 34.0 \\
\end{tabular}
\caption{\label{tab:table1}The average number of pages it takes to reach 90\% of the maximum recall and the number of pages it takes to reach the maximum recall. The maximum recall is considered the recall obtained after annotating all the pages of the book.}
\end{table}

\begin{figure}
\centering
\includegraphics[width=11cm]%,height=\textheight,keepaspectratio]
{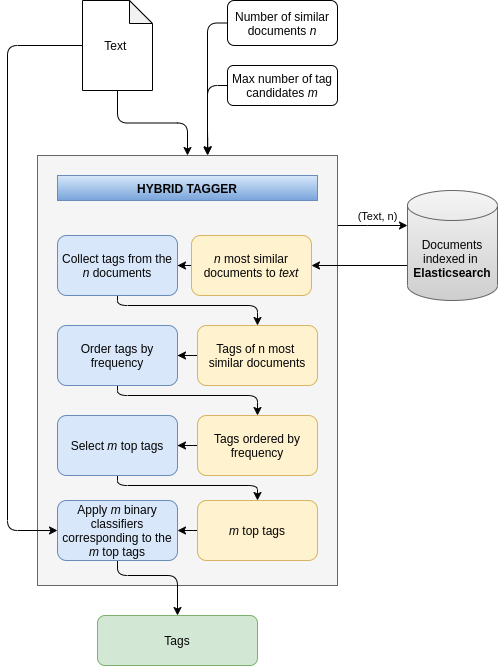}
\caption{The workflow of Hybrid Tagger tool used for predicting the keywords.}
\label{figure_3_hybrid_tagger}
\end{figure}

The average number of keywords added by the cataloguers is 8, meanwhile, the average number of keywords added by the keyword assignment model is 52. To retrieve fewer and more precise results, all the predicted keywords are first sorted by their frequency  (\textit{f = t/n}, where \textit{t} is the number of pages the keyword occurred in and n is the number of pages used) and only the keywords passing a configurable threshold are presented to the user.  The default threshold is set to 0.4 as it provides the highest F1 score on average and the number of keywords passing the threshold is most similar to the number of keywords added by the cataloguers on average (Figure \ref{figure_3_hybrid_tagger}). 

\section{Architecture}

The prototype consists of a back-end implemented as a RESTful Python application based on Django Rest Framework and an Angular front-end communicating with the backend via API endpoints. The prototype uses Celery for handling task queues and Redis as a message broker. Besides the main component, which is responsible for handling the subject indexing tool’s I/O, Kratt also includes Texta Toolkit containing the Hybrid Tagger used for predicting the labels, Texta MLP responsible for language extraction and preprocessing the data, and Elasticsearch for storing the data used by the Hybrid Tagger. All components, excluding Elasticsearch, are wrapped inside separate Docker containers to make them platform-independent and easily deployable. 

The tool can be used via graphical user interface or by passing the data directly to the API. 

\subsection{User Interface}

The books to annotate can be uploaded to the automatic subject indexing tool Kratt from the user’s computer or from an external resource by providing the URL of the resource. The user can additionally modify the number of randomly selected pages n used for further processing (by default 10). 

The subject indexing process takes about 1 minute, depending on the number of pages the user has selected. While the tool is processing the input, the user is displayed a progress chart with the information of the current step (e.g. “Detecting languages”). 

After the subject indexing tool has finished processing, the detected keywords passing the current threshold (by default = 0.4, but easily modifiable with a slider) are displayed along with the language distribution of the randomly selected pages. The user can deselect irrelevant keywords and then copy all selected keywords in MARC21 format to the clipboard.

\subsection{Workflow}

\begin{figure}
\centering
\includegraphics[width=\textwidth]%,height=\textheight,keepaspectratio]
{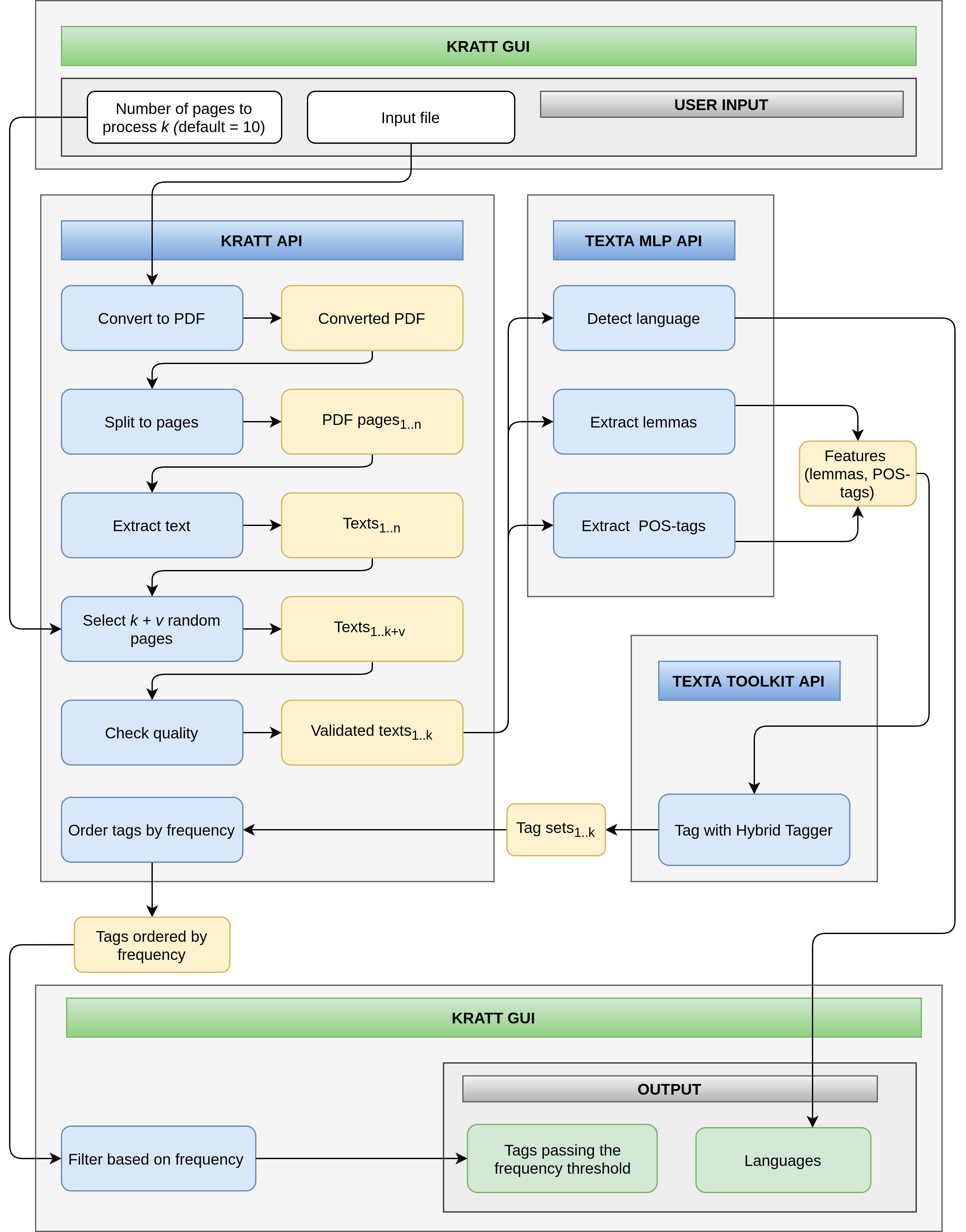}
\caption{The workflow of the automatic subject indexing tool Kratt.}
\label{figure_4_kratt_workflow}
\end{figure}

The workflow of Kratt consists of the following steps (Figure \ref{figure_4_kratt_workflow}):

\begin{enumerate}
\item The uploaded book is converted into PDF and divided into pages. 
\item Then \textit{n + 5} pages (the additional 5 pages are used as a buffer in case some of the selected pages do not pass quality control) are randomly selected and passed to the text extractor. 
\item The extracted plaintexts undergo quality control and \textit{n} texts passing the control are sent to Texta MLP. 
\item The lemmas and POS tags extracted with Texta MLP are then passed to the Hybrid Tagger, which predicts the keywords for each page. 
\item The keywords are then sorted by their frequency (\textit{f = t/n}, where \textit{t} is the number of pages the keyword occurred in and \textit{n} is the number of pages used). 
\item All keywords that exceed the configured threshold (by default 0.4) are presented to the user. 
\end{enumerate}

\section{Results and Discussion}

To evaluate the results, we applied the automatic subject indexing tool with default parameters to  315 new books subject indexed by the cataloguers during the 6 month period after finishing the prototype. The distribution of forms, genres and extent of the books was similar to the training data. However, the language distribution deviated from the original with only 59.0\% of the books in the test set being in Estonian - 27.5\% less compared to the training data. The other languages in the test set included English (28.9\%), Russian (7.3\%) and a small percentage of others (4.8\%). The books contained 1474 unique preferred terms of which 1222 (82.9\%) were present in the training data while developing the automatic subject indexing tool and 252 (17.1\%) were not. We excluded all the keywords not present in the training set while evaluating the results to get a better grasp of how well the model predicts keywords it should be able to predict and treat the rather large number of unseen labels as a separate problem. 

\begin{figure}[h]
\centering
\includegraphics[width=\textwidth]%,height=\textheight,keepaspectratio]
{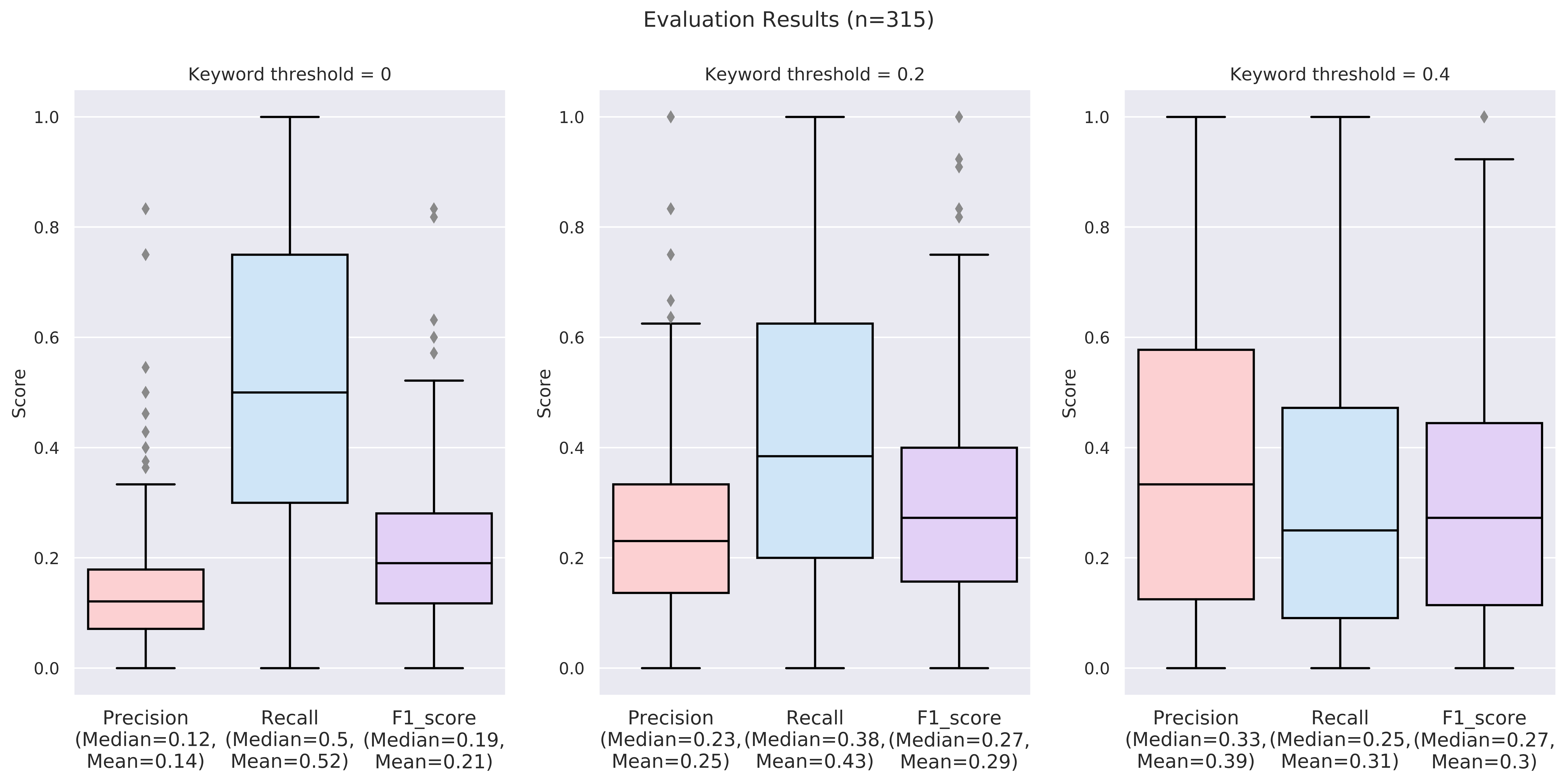}
\caption{The evaluation scores based on comparing the automatic subject indexing results of 315 books to the keywords added by the cataloguers. (1) Evaluation results without a threshold for keyword frequencies; (2) evaluation results with the keyword frequency threshold set to 0.2; (3) evaluation results with the keyword frequency threshold set to 0.4.}
\label{figure_5_eval_scores}
\end{figure}

We then compared the subject indices added by the cataloguers with the keywords predicted by the automatic subject indexing tool and calculated precision, recall and f1-score for each book. The overall scores were then calculated by averaging all the individual books’ scores. The distribution of resulting prediction scores on thresholds 0, 0.2, and 0.4 can be seen in Figure \ref{figure_5_eval_scores}. The best average f1-score was 0.3 and it was reached equally on thresholds 0.2 and 0.4, with the first providing higher average recall and the latter providing higher average precision.

Although the scores are rather underwhelming, there are several possible explanations and ways of improvement:

\begin{enumerate}
    \item The noise generated by assigning keywords to all pages might have more considerable effect than we originally anticipated. One possible solution for improvement is to try mapping only a relevant subset of the labels with each page and retrain the models.
    \item Although the number of examples per keyword increased remarkably after splitting each book into pages, the number of unique books linked with an average keyword stayed the same. This might have led to overfitting on the features tied to the content in the limited number of books. The best solution for this problem is to use more data, which could be accomplished by including the copyrighted books, which were not available during the development of the prototype due to legal complications. It is also worth exploring opportunities arising from large pretrained BERT\footnote{\url{https://github.com/google-research/bert}} models as they require less training data compared to the classical machine learning methods like logistic regression or support vector machines.
\end{enumerate}

In addition to the aforenamed explanations, we should also consider that the labels added by the cataloguers used for calculating the scores are subjective to some extent: even the labels chosen by two different cataloguers might not fully coincide. Furthermore, the subject indices are not usually 100\% correct or 100\% incorrect - there exists a grey area consisting of keywords, which are more or less adjacent to the topic, but considered excessive by the subject indexing methodology followed by the cataloguers. 

To get a better overview of the results regarding these aspects, we surveyed a small number of professional cataloguers and regular library users in addition to the automatic evaluation.

Five cataloguers working in the National Library of Estonia tested the tool with 20 books each. The cataloguers did not consider the results satisfactory and claimed that the results are rather causing extra confusion: if the prototype suggests a keyword originally not added by the cataloguer, it takes the cataloguer additional time to find out if the keyword proposed by the tool is actually relevant to the book's content and if it should be used in the subject index or not. These results are in alignment with a survey investigating the attitudes of German- and English speaking librarians towards automatic subject indexing. The findings of the survey showed that the librarians assess the quality of automatic subject indexing systems with scepticism and over 60\% of the respondents believed that machines will never be able to outperform humans in this task.\footnote{Keller. A. “Attitudes among German- and English-Speaking Librarians toward (Automatic) Subject Indexing”, Cataloging \& Classification Quarterly, 53:8 (2015), 895-904. DOI: 10.1080/01639374.2015.1061086}

As the methodology of cataloguing and subject indexing follows a strict set of rules,  the library workers may perceive the quality of the keywords differently from a regular user. To determine how the regular users perceive the results, we chose 10 random books from the test set and let 6 users evaluate the results based on the criteria presented in Table \ref{tab:table2}. To make sure that the randomly chosen books are representative of the test sample, we plotted the data points of the books on a figure representing the average prediction scores (f1-score, recall, precision) depending on the extent of the book. As most of the data points stay below the lines representing the average scores, we are confident that the chosen books did not distort the perception of the results of the automatic subject indexing tool in a favorable way. We presented three sets of subject indexing results for each book: 1) the original labels added by the cataloguers; 2) the labels added by the automatic subject indexing tool after using 10 pages of a book; and 3) the labels added by the automatic subject indexing tool after using all the pages of a book. The threshold was set to 0.2 to enhance recall rather than precision. The results grouped by different types of keywords are presented in Table \ref{tab:table3}. On average, all users evaluated the keywords added by the cataloguers higher than the keywords assigned by the  automatic subject indexing tool. However, there was a noticeable difference between different subcategories. While the topic keywords added by the cataloguers were considered better than the subject indexing tool’s results, Kratt outperformed cataloguers slightly with the prediction of genres. It is also worth mentioning that meanwhile the time keywords added by the cataloguers had a higher average score than the keywords suggested by the automatic subject indexing tool in the same category, the cataloguers added time keywords only to 2 books out of 10. Kratt predicted time keywords to 5-6 books out of 10 (depending on the number of pages used) and still received relatively high evaluation scores (4.5-4.6) from the users. As Kratt’s average results predicting the keywords is slightly over 4, we can conclude that the user’s perception of adequate keywords does not necessarily mirror the cataloguers’ perception. However, it should be noted that this specific evaluation process measured only how well the keywords were able to summarize the book, but this is not the only purpose of the keywords. They are also used for searching information from the library’s databases and while a couple of false positives or false negatives may not have a significant impact on the general overview of the book, they can negatively affect the user's experience while using the database. This also helps to explain the disparity between the users’ and cataloguers’ opinions as the latter group is used to considering all possible implications of the labelset.

\begin{table}[h]
\centering
\begin{tabular}{p{.1\linewidth}|p{.8\linewidth}}
\textbf{Rating} & \textbf{Description} \\\hline
1 & The keywords are irrelevant and do not represent the content of the book.\\
2 & Some of the keywords are accurate, but the larger majority does not represent the content of the book \\
3 & Fair amount of the keywords are relevant to the topics covered in the book, but there exists a sufficiently significant amount of irrelevant keywords to cause confusion. \\
4 & Most of the keywords are relevant, but a few are not. \\
5 & The keywords give a decent overview of the book and are relevant to the topics covered in the book. \\
\end{tabular}
\caption{\label{tab:table2}The description of the ratings used for surveying the users about the keywords added by a) the cataloguers b) the automatic subject indexing tool Kratt.}
\end{table}

\begin{table}[h]
\centering
\begin{tabular}{p{.20\linewidth}|p{.1\linewidth}|p{.14\linewidth}|p{.1\linewidth}|p{.1\linewidth}|p{.14\linewidth}}
 & \textbf{Topic} & \textbf{Location} & \textbf{Time} & \textbf{Genre} & \textbf{ All keywords} \\\hline
\textbf{Cataloguer}& 4.59 & 4.69 & 5 & 4.28 & 4.64\\
\textbf{Kratt / 10 pages} & 3.43 & 4.32 & 4.5 & 4.31 & 4.14 \\
\textbf{Kratt/ All pages} & 3.98 & 4.38 & 4.6 & 4.33 & 4.32 \\
\end{tabular}
\caption{\label{tab:table3}The average user ratings of different types of keywords added by a) the cataloguers b) the automatic subject indexing tool Kratt.}
\end{table}

As one of the goals of developing the automatic subject indexing tool is to save resources, we measured the time Kratt spends for subject indexing a book and compared it with the performance of an average cataloguer. It currently takes Kratt about 1 minute to predict keywords for a book of any size with the default settings. There is no additional time cost for longer books as only 10 random pages are used for predictions (by default) and the time consumption of processes like converting and splitting the file is negligible. It should also be noted that increasing the training data and the set of keywords does not affect the time consumption as the prediction time of Hybrid Tagger does not depend on the number of possible targets. As it takes an average cataloguer about 15 minutes to subject index a book, the speed of automatic subject indexing tool outperforms human specialists. Furthermore, the automatic subject indexing tool does not need to rest and can work 24 hours a day, 7 days a week.

As one of the goals of developing the automatic subject indexing tool is to save resources, we measured the time Kratt spends for subject indexing a book and compared it with the performance of an average cataloguer. It currently takes Kratt about 1 minute to predict keywords for a book of any size with the default settings. There is no additional time cost for longer books as only 10 random pages are used for predictions (by default) and the time consumption of processes like converting and splitting the file is negligible. It should also be noted that increasing the training data and the set of keywords does not affect the time consumption as the prediction time of Hybrid Tagger does not depend on the number of possible targets. As it takes an average cataloguer about 15 minutes to subject index a book, the speed of automatic subject indexing tool outperforms human specialists. Furthermore, the automatic subject indexing tool does not need to rest and can work 24 hours a day, 7 days a week.

\section{Conclusion}

The goal for developing the prototype of an automatic subject indexing tool Kratt was to test whether the automation of the process would be possible, if it might save time and money, if it would help to improve the quality and if Kratt could be integrated into the daily cataloguing workflows. Our results demonstrated that while the automation process is possible and more time efficient than manual cataloguing, the quality of the predicted subjects is currently not sufficient for integrating the automated process into the library’s daily workflows. However, we do not rule out the possibility of including it in the future, if the proposed methods of enhancing the quality of the models prove to be successful.

\section{Funding}
The project of developing the prototype of an automatic subject indexing tool was supported and financed from the European Regional Development Fund (ERDF). The work described in this paper has also been supported by the language technology research and development program ”Estonian Language Technology 2018–2027” of the Ministry of Education and Research under grant EKTR3, by European Union’s Horizon 2020 research and innovation programme under grant agreement No 825153, project EMBEDDIA (Cross-Lingual  Embeddings for LessRepresented Languages in European News Media) and Enterprise Estonia project No. EU48684, Research Project No. 1.11 (Deep neural models and cross-lingual embeddings in TEXTA Toolkit).

\section{Data Availability Statement}
The processed data used for developing the prototype described in this study are available in DIGAR at \url{http://data.digar.ee/#kratt}. These data were derived from the following resources available in the public domain: \url{https://www.digar.ee/arhiiv/en} (book files), \url{https://erb.nlib.ee/?lang=eng} (metadata) 

\theendnotes

\end{document}